\documentclass{article}

\usepackage[preprint,nonatbib]{nips_2018}

\usepackage[utf8]{inputenc} 
\usepackage[T1]{fontenc}    
\usepackage[hidelinks]{hyperref}       
\usepackage{url}            
\usepackage{booktabs}       
\usepackage{amsmath,amssymb}
\usepackage{mathtools, bm}
\usepackage{nicefrac}       
\usepackage{microtype}      

\newcommand{\rd}[1][]{\mathrm{d}#1}

\usepackage[round]{natbib}

\title{Variational Bridge Constructs for Approximate Gaussian Process Regression}

\author{
  Wil O C Ward \\ 
  Department of Computer Science\\
  Univesity of Sheffield\\
  Sheffield, UK\\
  \texttt{w.ward@sheffield.ac.uk}
  \And
  Mauricio A \'Alvarez\\
  Department of Computer Science\\
  University of Sheffield\\
  Sheffield, UK\\
  \texttt{mauricio.alvarez@sheffield.ac.uk}
}

\begin{document}
\maketitle

\begin{abstract}
This paper introduces a method to approximate Gaussian process regression by representing the problem as a stochastic differential equation and using variational inference to approximate solutions. The approximations are compared with full GP regression and generated paths are demonstrated to be indistinguishable from GP samples. We show that the approach extends easily to non-linear dynamics and discuss extensions to which the approach can be easily applied.
\end{abstract}

\section{Introduction}
Gaussian process (GP) regression is an effective tool for inferring the nature of a function given a small number of possibly noisy observations. However, in the case where these signals interact non-linearly, for example in non-linear multi-output GPs and latent force models (LFMs) \citep{Hartikainen2012}; or where the likelihoods are non-Gaussian \citep{Hensman2015}, approximation methods are necessary. Challenges lie in the effective estimation of the posterior in the case where it is intractible, as well as scalability for large datasets.

In the case of the former, there are a number of approaches including  variational inference (VI). VI can also be utilised with sparse representations of the Gaussian approximation to deal with large datasets \citep{Hensman2013}. Alternatively, the GP can be represented as a (Markovian) stochastic differential equation (SDE) and inferred sequentially in linear time \citep{Hartikainen2010}. In more recent work, this approach has been adapted for non-Gaussian likelihoods \citep{Nickisch2018}.

The representation of GPs as SDEs has been used with covariance assumptions on, for example, smooth and (quasi-)periodic systems \citep{Hartikainen2010,Solin2013}. In these instances, the posterior can be inferred using Kalman filtering and smoothing \citep{Sarkka2013}; non-linear approximations thereof have been used for models with non-linear interaction, such as in the case of non-linear LFMs represented as SDEs \citep{Hartikainen2011}. An alternative approach to approximating non-linear SDEs was proposed by \citet{Ryder2018}, constructing Brownian bridges conditioned on observations and using a variational distribution to represent the posterior solution to the SDE.

In this paper, we utilise these \emph{variational bridges} to approximate Gaussian process regression in its SDE form, comparing the approximation against full GP regression. We demonstrate that the paths generated are effective approximations of a true GP, using maximum mean discrepancy metrics to highlight that they are statistically indistinguishable from GP samples. We also show that the proposed approach is capable of inferring a latent GP observed through a non-linear mapping, and discuss how this can be extended to more complicated non-linear problems.

\section{Sequential Gaussian Process Regression}\label{sec:seq_regression}
Consider a 1-D (e.g. temporal) zero-mean Gaussian process, $f(t) \sim \mathcal{GP}(0,k(t,t'))$, with covariance function $k(t,t')$. Full batch GP regression involves calculating the covariances between input values within and between some observed set and some test set, and involves the inversion of a potentially large covariance matrix. However, the interpretation of the regression problem as a stochastic differential equation allows for solution in a sequential manner \citep{Hartikainen2010}.

A Gaussian process can be described as a summation of differential operators driven by a stochastic white-noise process \citep{Sarkka2019}, in particular an It\^o SDE in companion form: $\rd{\bm{f}(t)} = \mathbf{F}\bm{f}(t)\rd{t} + \mathbf{L}\rd{\bm{\beta}(t)}$. The augmented state is the latent function and its derivatives,
\begin{align*}
  \bm{f}(t) = \begin{bmatrix} f(t) & \frac{\rd}{\rd{t}}f(t) & \cdots & \frac{\rd^{m-1}}{\rd{t^{m-1}}}f(t)\end{bmatrix}^\text{T}.
\end{align*}
The SDE is driven by a white-noise process, which is itself a Gaussian process by virtue of it being the derivative of Brownian motion so defined with covariance $k_w(t,t') = q\delta_{t,t'}$.

The nature of the covariance function of the GP prior dictates the form of $\mathbf{F}$ and $\mathbf{L}$, the structure of which is linked to the dimension of the latent state $\bm{f}$ and of the white noise process. For example, the family of Mat\'ern covariance functions, particularly those with half-integer smoothness $\nu=p+1/2$ are of interest. In the latter case, the covariance functions are differentiable $p$ times, restricting the dimension of $\mathbf{F}$ to $p\times p$. Likewise, the spectral density, $q$ of the white noise process is a scalar and so $\mathbf{L}$ is $p\times 1$.

For the regression problem, with latent function can be mapped as $f(t) = [1\ 0\ \ldots\ 0]\bm{f}(t)$, and the discretised system is a state-space model and can be solved in linear time with the Kalman filter and Rauch-Tung-Streibel smoother \citep{Hartikainen2010}.

\section{Variational Bridges for Gaussian Process Regression}
Consider an It\^o process, driven by the SDE $\rd{\bm{f}}(t) = \bm{g}(\bm{f}(t),\theta)\rd{t} + \bm{c}(\bm{f}(t),\theta)\rd{\beta}$, where $\bm{g}$ denotes the drift term and $\bm{c}^2(\cdot)\triangleq\bm{c}(\cdot)\bm{c}(\cdot)^\text{T}$ is the diffusion matrix of the SDE. Solutions to the SDE may be intractible and must be approximated in some way, most commonly with iterative solvers. One example is the Euler-Maruyama method, which describes \emph{transition densities} as:
\begin{align}
  f_{k+1}\,|\, f_k, \theta \sim \mathcal{N}\!\left(f_k + \Delta_t\bm{g}(f_k,\theta), \bm{c}^2(f_k, \theta)\Delta_t\right),\label{eq:euler-maruyama}
\end{align}
where $f_k \triangleq \bm{f}(t_k)$, the evaluation of $\bm{f}(\cdot)$ at some discrete time-step $t_k$; and $\Delta_t$ is the step interval, $t_{k} - t_{k-1}$. The GP model is defined such that $\bm{g} \triangleq \mathbf{F}_\theta\bm{f}(t)$; $\bm{c} \triangleq \mathbf{L}\sqrt{q}$; and $\theta$ is the set of hyperparameters of the covariance function. Given some observations $\mathbf{y}$ at times $[\tau_j]_{j=0}^N$, where the observation model is defined $y_j = h(\bm{f}(\tau_j)) + \varepsilon_j$, conditioning the SDE on the observations gives a Brownian bridge construct with diffusion defined by the GP prior.

In \citet{Ryder2018}, the authors use black-box variational approach to designing Brownian bridge constructs. We utilise this approach on the SDE representation of the Gaussian process as described in Section~\ref{sec:seq_regression}, conditioning the diffusion process on observations. The variational approximation of this path, $q_\phi(\bm{f}\,|\,\theta)$, has the form
\begin{align}
  q_\phi(\bm{f}\,|\,\theta) = \prod^T_{k=0}\mathcal{N}\!\left(f_{k+1}-f_k\,|\,\tilde{\bm{g}}(f_k,\mathbf{y},\theta,t_k,\phi)\Delta_t, \tilde{\bm{c}}^2(f_k,\mathbf{y},\theta,t_k,\phi)\Delta_t\right),\label{eq:var_dist}
\end{align}
where $\tilde{\bm{g}}$ and $\tilde{\bm{c}}^2$ are the drift and diffusion terms of the variational path.

As an iterative model, driven by Gaussian noise, the variational distribution can also be represented as an implicit generative function $\bm{f} = \mathfrak{g}(\epsilon, \theta, \phi)$, the result of iteratively solving the reparameterised terms in the product of \eqref{eq:var_dist}.

The evidence lower bound (ELBO) of the variational parameters, $\mathcal{L}(\phi)$ is maximised using $n$ Monte Carlo realisations of the variational paths and hyperparameters:
\begin{align}
  \hat{\mathcal{L}}(\phi) = \frac{1}{n}\sum^n_{i=1}\log\frac{p(\theta^{(i)})p(\bm{f}^{(i)}\,|\,\theta^{(i)})p(\mathbf{y}\,|\,\bm{f}^{(i)}, \theta^{(i)})}{q_\phi(\theta^{(i)})q_\phi(\bm{f}^{(i)}\,|\,\theta^{(i)})},
\end{align}
where $\bm{f}^{(i)} = \mathfrak{g}(\epsilon, \theta^{(i)}, \phi_f)$, the implicit model parameterised by a recurrent neural network (RNN).

The variational distribution of the kernel hyperparameters used in the experiments is a mean-field Gaussian approximation, where $q_\phi(\theta) = \prod\mathcal{N}(\theta|\mu_i,s_i)$. The variational parameters are thus the mean and covariance of the Gaussians in the product term: $\phi_\theta = \{(\mu_i,s_i)\,|\,i=1,\ldots,n_\theta\}$. Likewise, the variational distribution of the latent function $\bm{f}$ uses an Euler-Maruyama discretisation and constructs Brownian bridges conditioned on the observations, represented by an RNN. The variational parameters $\phi_f$ in this case are the weights of the RNN.

The generative function parameterised with an RNN represents the terms $\tilde{\bm{g}}$ and $\tilde{\bm{c}}$, similar to \citet{Ryder2018}; the implementation of the RNN cells uses the same set up and assumptions. The inference is then performed by maximising the ELBO given the posterior of the SDE as defined using the Euler-Maruyama discretisation in \eqref{eq:euler-maruyama}.

\section{Experiments}

\begin{figure}[t!]
  \centering
  \includegraphics[width=0.9\textwidth]{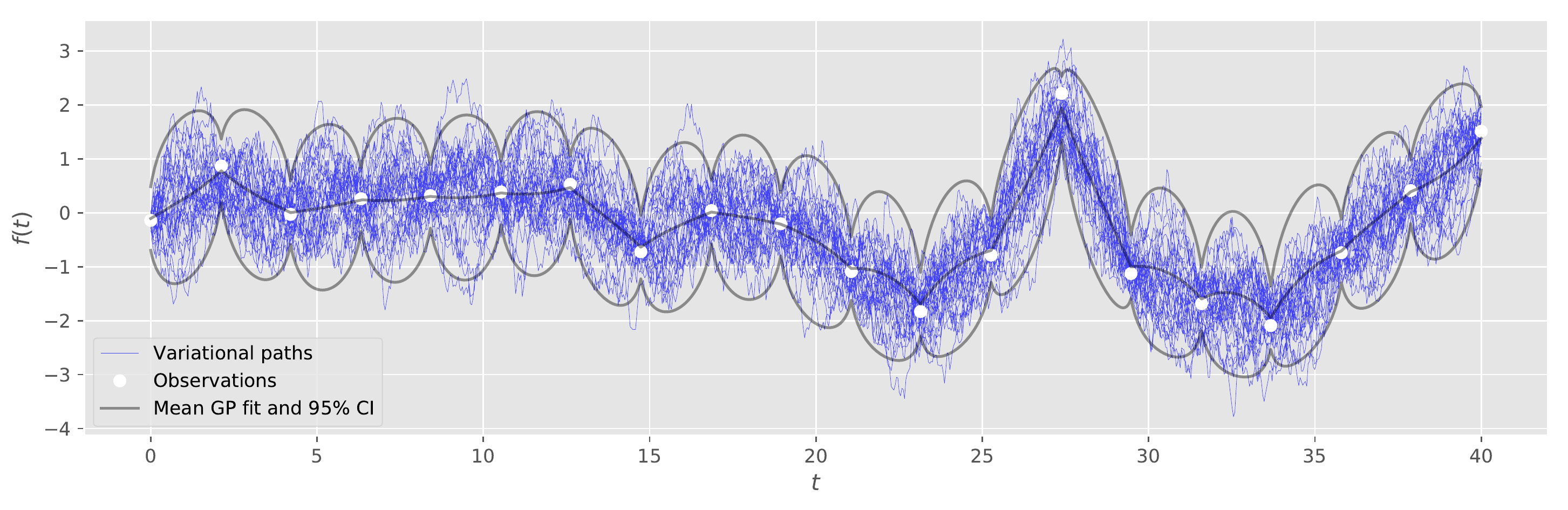}
  \vspace{-5pt}
  \caption{Variational paths after training for 25k epochs in comparison with a full GP regression}
  \label{fig:expgp}
  \vspace{-10pt}
\end{figure}

\subsection{GP Regression with Exponential Covariance}
In this example, Gaussian process regression is performed with a zero-mean prior and the exponential covariance function: $k(t,t') = \sigma^2_k\exp(-\lambda|t-t'|)$. A random path is generated, with discrete observations sampled at uniform intervals with additive Gaussian noise with variance $\sigma_y^2=0.1$.

The underlying white-noise driven SDE of the GP with an exponential covariance is defined as
\begin{align*}
  \rd{f(t)} = -\lambda f(t) + \rd{\beta(t)},
\end{align*}
with the spectral density of the white-noise process, $q=2\sigma^2_k\lambda$. The hyperparameters, $\theta=\{\lambda,\sigma^2_k\}$ were approximated with a mean-field Gaussian variational distribution, and updated iteratively with the RNN weights during training.

The 30 variational paths generated from 25,000 epochs of training are shown in \figurename{}~\ref{fig:expgp}, along with the sample moments that approximate our variational GP. Overlaid are the moments of a full GP regression using the exponential kernel. We observe visually that mean and credible interval of the GP maps represents a similar fit to the paths.

\subsection{Model Criticism}
To indicate the reliability of the variational approximation of a Gaussian process, we compare the variational paths generated with samples from the full GP in \figurename{}~\ref{fig:expgp}. We map the corresponding paths and samples into a reproducing kernel Hilbert space (RKHS), using a Gaussian kernel, and apply a two-sample test using maximum mean discrepancy (MMD) to provide a metric of the similarity of the distributions generated by the paths and from which the samples are drawn \citep{Gretton2012}.

\begin{table}[t!]
\centering
\caption{Unbiased MMD$^2$ estimations comparing variational paths with samples from a GP for different training epochs and number of observations. Also shown are the respective thresholds indicating 95\% rejection}\label{tab:mmdvals}
\begin{tabular}{c | c c c c c c | c}
  Epoch & 10 & 100 & 500 & 1 000 & 2 500 & 25 000 & Threshold\\
  \midrule
  $T=6$ & 0.11106 & 0.12670 & 0.05960 & 0.04843 & 0.05559 & -- & 0.03713\\
  $T=20$ & 0.27305 & 0.11473 & 0.06539 & 0.06962 & 0.04714 & \textbf{0.03157} & 0.03368
\end{tabular}
\end{table}

In Table~\ref{tab:mmdvals}, the MMD$^2$ value of the variational paths and samples from the full GP fit are shown against the training epoch. As the training time increases, we observe the discrepancy between variational approximation and GP fit decreases, and after 25,000 epochs, the variational paths are indistinguishable from samples from the GP. The results also demonstrate that the similarity between variational paths and GP samples is present even with a small number of observations, due to the conditioned Brownian bridge encoding the GP prior.

\begin{figure}[t!]
  \centering
  \includegraphics[width=0.9\textwidth]{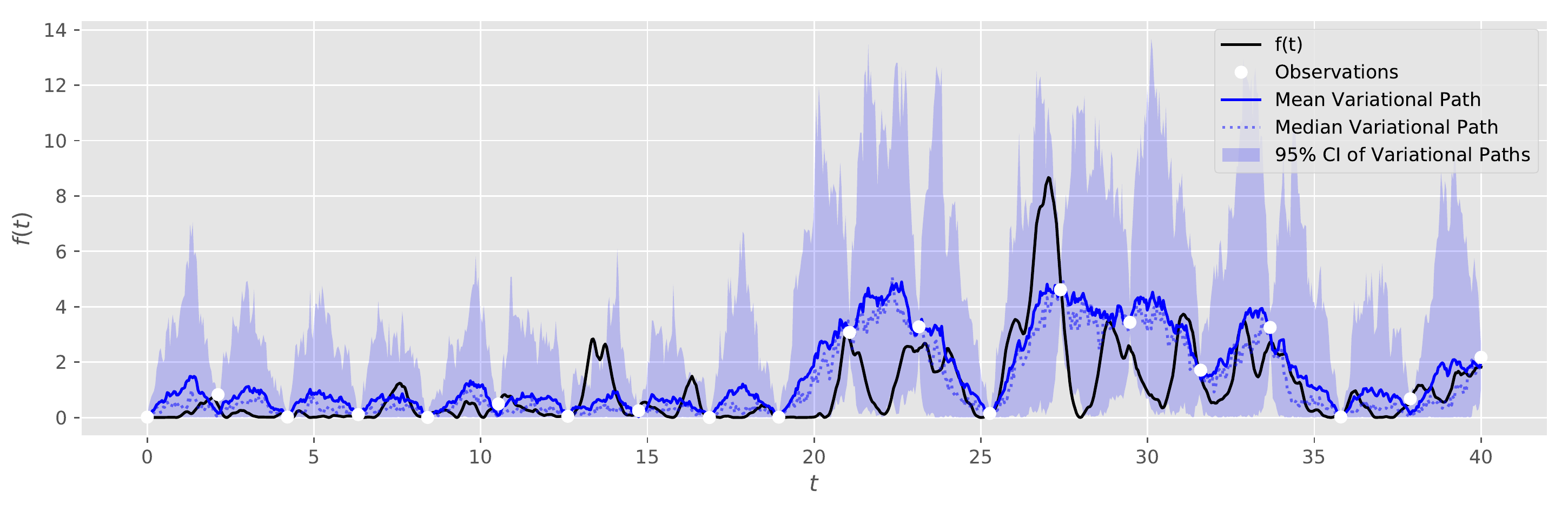}
  \vspace{-5pt}
  \caption{Summary statistics of variational paths with GP prior and non-linear likelihood}
  \label{fig:exp_nonlin}
  \vspace{-10pt}
\end{figure}

\subsection{Non-Linear Likelihood}
To demonstrate the efficacy in cases where straightforward batch or sequential GP regression is intractible, we perform a similar regression problem on a latent function for which the Gaussian process is related non-linearly.

Consider a latent function with GP prior, $u$, noisly observed through a non-linear mapping in $f$:
\begin{align*}
  f(t) = u^2(t), \qquad y_k = f(\tau_k) + \varepsilon_k.
\end{align*}
It's evident here that exact Gaussian process regression on $\mathbf{y}$ is not easily performed directly. In this experiment, we fit variational bridges to $u(t)$ using the approach described in this paper. Again, the exponential covariance was assumed in the prior over our GP in the regression problem.

The mean and 95\% confidence interval of the variational paths approximating $f(t)$ are plotted in \figurename{}~\ref{fig:exp_nonlin}, demonstrating the efficacy of the variational approximation for regression over non-linear observations. We can see this in particular, where the confidence interval captures the trough between the observations at $t\approx21$ and $t\approx23$, as well as the peak at $t\approx27$.

\section{Discussion}
We demonstrate that we can approximate GP regression using variational bridge constructs by encoding the GP assumptions of a signal as an SDE. The generated variational paths are shown to be indistinguishable from a GP obtained with full batch GP regression, indicating that the approach can produce valid GP samples. While the approach is not intended as a replacement to batch GP regression directly, we show that it is an effective tool for approximating functions where the likelihoods make the regression intractible.

The approach can be futher extended to handle non-linear problems for which there are GP priors involved, utilising the advantages of the variational bridges to solve non-linear SDEs. For example non-linear latent force models \citep{Alvarez2013}, including those problems with non-Gaussian likelihoods, could be approximated with little adjustment to the underlying model. The approach can also make use of past work on using priors with other covariance functions with sequential GP inference, such as periodic and RBF covariances \citep{Solin2013, Hartikainen2010}.

\paragraph{Acknowledgement} The authors would like to thank Dennis Prangle and Dougal Sutherland for their input and advice. WW and MAA have been financed by the Engineering and Physical Research Council (EPSRC) Research Project EP/N014162/1.

\bibliographystyle{abbrvnat}
\bibliography{manuscript}
\end{document}